%% file: main.tex
\documentclass[conference]{IEEEtran}
\IEEEoverridecommandlockouts
\usepackage{cite}
\usepackage{algorithmic}
\usepackage{graphicx}
\usepackage{textcomp}
\usepackage{xcolor}
\def\BibTeX{{\rm B\kern-.05em{\sc i\kern-.025em b}\kern-.08em
    T\kern-.1667em\lower.7ex\hbox{E}\kern-.125emX}}

\usepackage{amsmath,amssymb,amsfonts}
\usepackage{graphicx}
\usepackage{booktabs}
\usepackage{makecell}
\usepackage{subcaption}
\usepackage{fancyhdr}
\usepackage{hyperref}

\usepackage[colorinlistoftodos,prependcaption,textsize=tiny]{todonotes}
\usepackage{xargs}                      
\newcommandx{\burak}[2][1=]{\todo[linecolor=green,backgroundcolor=green!25,bordercolor=green,#1]{#2}}

\title{Bit Distribution Study and Implementation of Spatial Quality Map in the JPEG-AI Standardization}
\author{
	\IEEEauthorblockN{
		Panqi Jia\IEEEauthorrefmark{1}\IEEEauthorrefmark{3}, Jue Mao\IEEEauthorrefmark{3}, Esin Koyuncu\IEEEauthorrefmark{1}\IEEEauthorrefmark{3},A. Burakhan Koyuncu\IEEEauthorrefmark{2}\IEEEauthorrefmark{3},\\ Timofey Solovyev\IEEEauthorrefmark{3}, Alexander Karabutov\IEEEauthorrefmark{3},
		Yin Zhao\IEEEauthorrefmark{3}, Elena Alshina\IEEEauthorrefmark{3}, André Kaup\IEEEauthorrefmark{1}}\\
	\IEEEauthorblockA{\IEEEauthorrefmark{1}Multimedia Communications and Signal Processing, Friedrich-Alexander University Erlangen-Nürnberg, Germany}
	\IEEEauthorblockA{\IEEEauthorrefmark{2}Chair of Media Technology, Technical University of Munich, Germany}
	\IEEEauthorblockA{\IEEEauthorrefmark{3}Huawei Technologies}
}

\begin{document}
\maketitle
\input{abstract}
\input{sections/introduction}

\label{sec:intro}

\input{sections/BDM}


\input{sections/QPM}

\input{sections/Conlusion}

\bibliographystyle{IEEEtran}
\bibliography{References}
\end{document}

%% file: abstract.tex
\begin{abstract}
Currently, there is a high demand for neural network-based image compression codecs. These codecs employ non-linear transforms to create compact bit representations and facilitate faster coding speeds on devices compared to the hand-crafted transforms used in classical frameworks. The scientific and industrial communities are highly interested in these properties, leading to the standardization effort of JPEG-AI.  The JPEG-AI verification model has been released and is currently under development for standardization. Utilizing neural networks, it can outperform the classic codec VVC intra by over 10\% BD-rate operating at base operation point. Researchers attribute this success to the flexible bit distribution in the spatial domain, in contrast to VVC intra's anchor that is generated with a constant quality point. However, our study reveals that VVC intra displays a more adaptable bit distribution structure through the implementation of various block sizes. As a result of our observations, we have proposed a spatial bit allocation method to optimize the JPEG-AI verification model's bit distribution and enhance the visual quality. Furthermore, by applying the VVC bit distribution strategy, the objective performance of JPEG-AI verification mode can be further improved, resulting in a maximum gain of 0.45 dB in PSNR-Y.

\end{abstract}

%% file: sections/introduction.tex
\section{Introduction}
\noindent Image compression codecs aim to achieve high quality reconstruction through efficient bitstream utilization. Currently, these codecs can be categorized into two types: classical codecs and neural network-based codecs.

The classical codes contain JPEG \cite{125072}, JPEG2000 \cite{952804}, HEVC \cite{6316136}, BPG \cite{bellard2015bpg}, which is based on HEVC intra, and VVC \cite{9301847}. Classical image compression codecs usually process the image block by block, and they have intra prediction, quantization, arithmetic coding, and de-blocking filters. 

In recent times, diverse compression algorithms based on neural networks (NN) have been explored to enhance compression capabilities~\cite{Ball2017EndtoendOI,Ball2018VariationalIC,Minnen2018JointAA,cheng2020learned,koyuncu2021parallelized,guo2021causal,qian2021entroformer,koyuncu2022contextformer,he2022elic,koyuncu2023efficient,liu2023learned}. Those algorithms utilise non-linear transform coding~\cite{balle2020nonlinear} through an autoencoder-based architecture employing non-linear transforms. This process facilitates the mapping of the input image to a concise latent representation. The transform coefficients are acquired end-to-end with an accompanying entropy model that approximates the entropy of the latent variables. Previous proposals~\cite{Ball2017EndtoendOI,Ball2018VariationalIC,Minnen2018JointAA,cheng2020learned,koyuncu2021parallelized} achieve compression performance comparable to that of classical codecs, including JPEG \cite{125072}, JPEG2000 \cite{952804}, and BPG \cite{bellard2015bpg}. More recent approaches~\cite{guo2021causal,qian2021entroformer,koyuncu2022contextformer,he2022elic,koyuncu2023efficient,liu2023learned} introduce transform layers with sophisticated activation functions~\cite{he2022elic,liu2023learned}, and entropy model techniques with attention modules~\cite{guo2021causal,qian2021entroformer,koyuncu2022contextformer,koyuncu2023efficient,liu2023learned}. Those studies outperform VVC in terms of compression performance~\cite{9301847}, and also achieve the same runtime complexity as traditional codecs~\cite{he2022elic,koyuncu2023efficient,liu2023learned}. However, they can only attain various levels of compression by training distinct models for each rate point and are unable to offer continuous variable rates through a single model during evaluation. 

To accomplish continuous variable rate coding, a variety of revisions have been suggested for NN-based compression algorithms~\cite{9522770,choi2019variable,song2021variable,Cui2020GVAEAC,cui2021asymmetric}. F. Brand et al.~\cite{9522770} proposed a pyramidal NN configuration with latent masking for variable rate coding. In this approach, the mask for each layer is fine-tuned to optimize the encoding process. Choi et al.~\cite{choi2019variable} and Song et al.~\cite{song2021variable} have presented the concept of using conditional transform layers, where transforms are tailored in accordance with the target rate. In addition, the method proposed in~\cite{song2021variable} allows the allocation of additional bits exclusively to the region of interest (ROI), thereby amplifying the encoding quality of the chosen location. Alternatively, the~\cite{Cui2020GVAEAC,cui2021asymmetric} model proposes a so-called \textit{gain unit}, which allows for direct adjustment of latent variables without the need to modify the transform layers. This trainable matrix integrates various gain vectors, supplying channel-specific quantization maps and enabling continuous variable rates through gain vector extrapolation or interpolation.

\begin{figure}
    \centering
    \setkeys{Gin}{width=0.48\linewidth} 
\subfloat[The original image \label{fig:BDM_05_i}]{\includegraphics{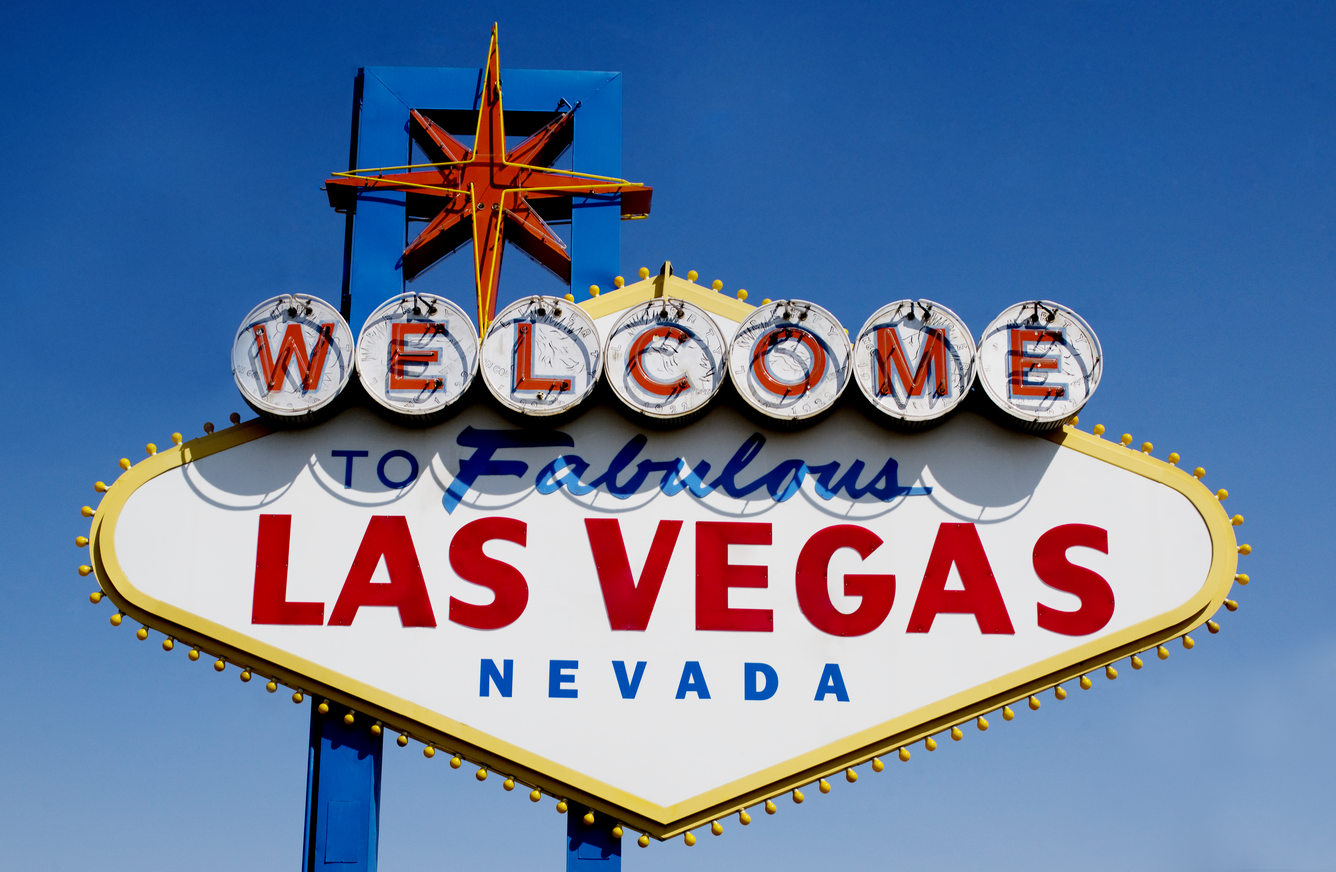} }\hfil
\subfloat[original BDM of VVC \label{fig:BDM_05_b}]{\includegraphics{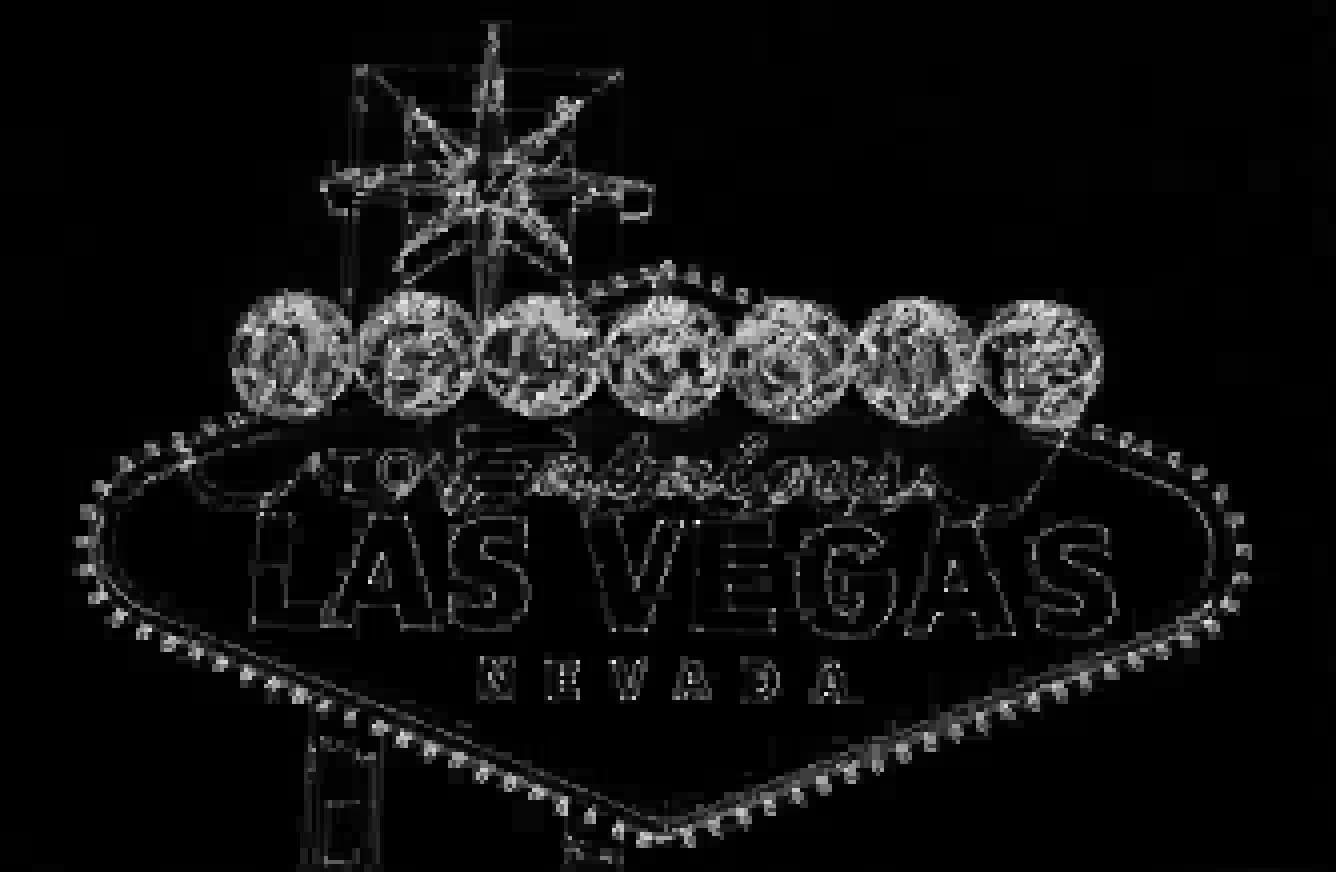} }\hfil
\subfloat[Regrouped $16\times 16$ block BDM of VVC \label{fig:BDM_05_c}]{\includegraphics{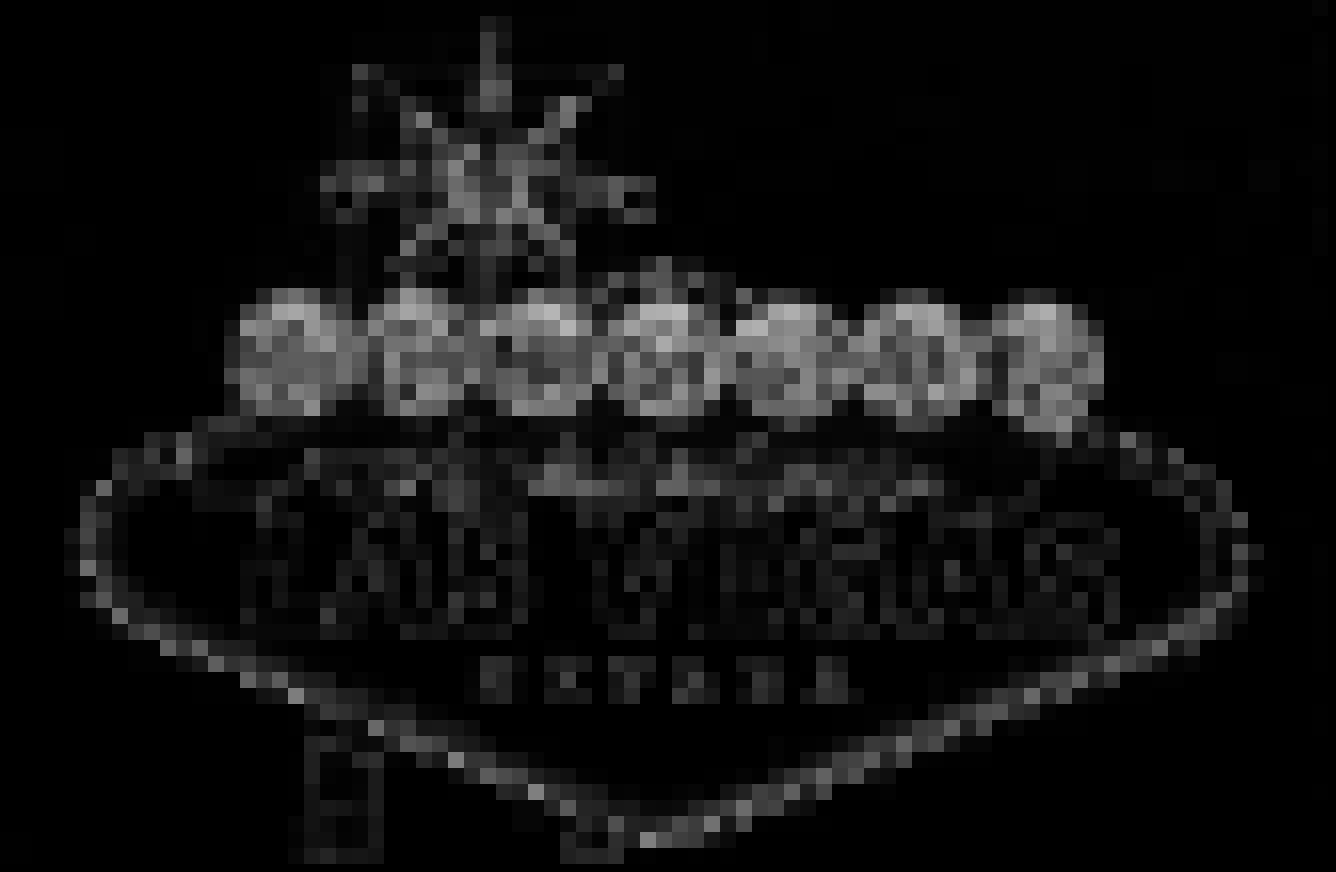} }\hfil
\subfloat[BDM of JPEG-AI VM \label{fig:BDM_05_a}]{\includegraphics{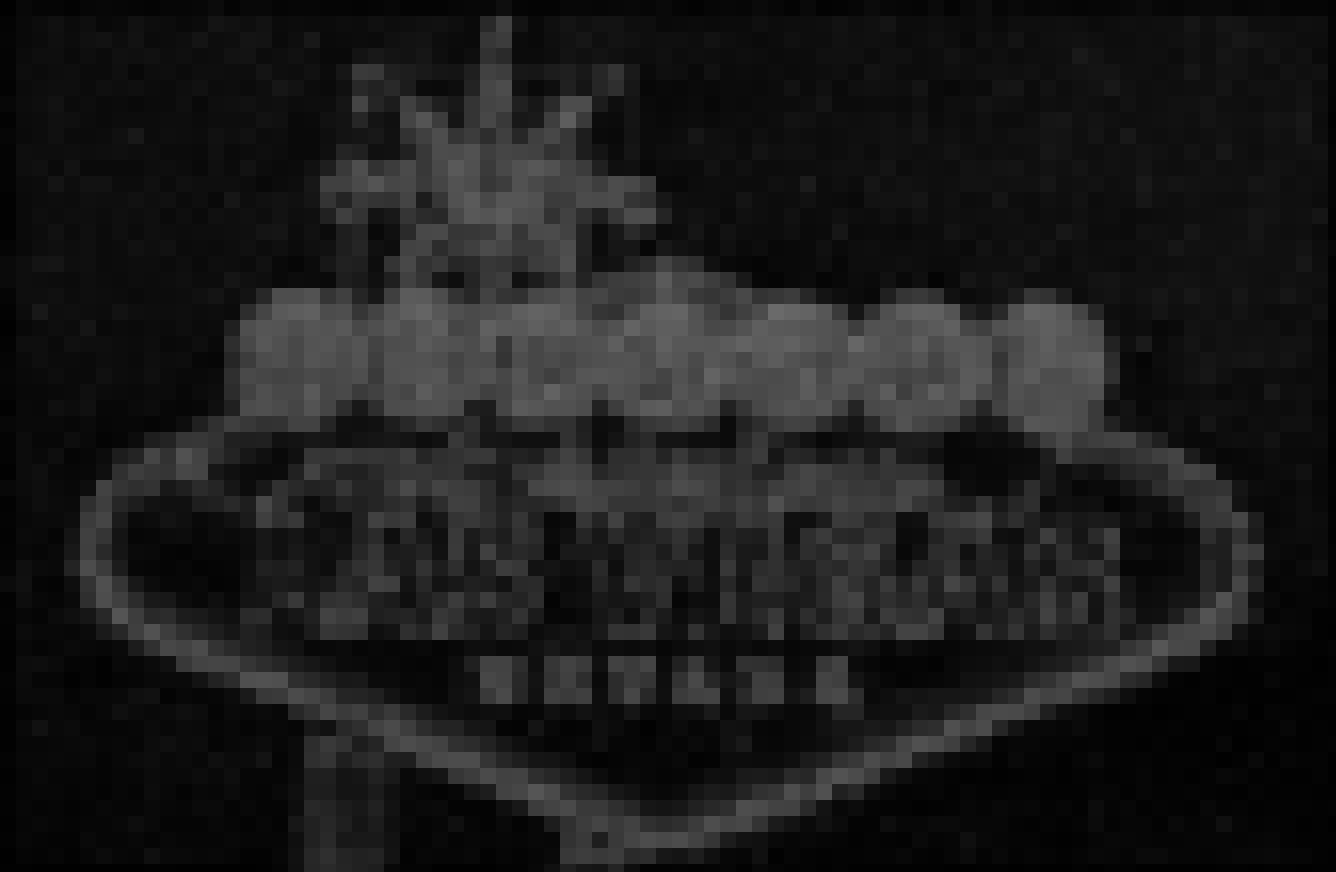} }\hfil

\caption{Luminance BDM of the 5th image at 0.75 bpp}
\label{fig:BDM_05}

\end{figure}

\begin{table*}[t]
	\centering
	\footnotesize
	\begin{tabular}{c|ccc|ccc}
		\toprule
		\textbf{metrics}&\textbf{IMG05 0.75bpp}&\textbf{IMG34 0.75bpp}&\textbf{IMG43 0.75bpp}&\textbf{IMG12 0.25bpp}&\textbf{IMG12 0.5bpp}&\textbf{IMG32 0.75bpp}\\
		\midrule
		\makecell[t]{PSNRY of JPEG-AI VM \\PSNRY of VVC\\BDM Var. of JPEG-AI VM\\BDM Var. of VVC}
		
		&\makecell[t]{39.80\\46.73\\0.0074\\0.0164}
		
		&\makecell[t]{33.49\\38.33\\0.013\\0.0334}
		
		&\makecell[t]{41.61\\46.61\\0.0083\\0.0212}
		
		&\makecell[t]{33.24\\32.67\\0.004\\0.0041}
		
		&\makecell[t]{37.12\\36.79\\0.0053\\0.0064}
		
		&\makecell[t]{35.99\\34.93\\0.008\\0.0141}
		\\
		\bottomrule
	\end{tabular}
	\caption{PSNR Y (in dB) and BDM Var. of JPEG-AI VM and VVC, the selected images at specific bpp shows the biggest gap in PSNRY}
	\label{tab1}
\end{table*}

Recent efforts to standardise JPEG-AI~\cite{ascenso2023jpeg} have investigated the effectiveness of NN-based encodings, highlighting their impressive compression capabilities. Notably, a verification model (VM)~3.4 has been launched, underpinned by the conditional colour separation (CCS) framework suggested in~\cite{jia2022learningbased}. VM~3.4 provides two operational points to cater to diverse complexity needs. The high operation point (HOP) provides a notably enhanced compression performance when compared to VVC~\cite{9301847}. Conversely, the base operation point (BOP) yields only moderate performance improvements despite significantly lower model and runtime complexities. Additionally, VM~3.4 utilises a gain unit similar to that of~\cite{Cui2020GVAEAC} to assist in continuous variable rate coding. However, the gain unit is only capable of providing a quantization map on a channel-by-channel basis to alter the overall bits per pixel (bpp). In this work, a spatial bit distribution method is proposed to support different bpp in different regions of the image. And it helps to improve both the subjective and objective quality of the JPEG-AI VM.

%% file: sections/BDM.tex
\section{Spatial Bit Distribution Analysis between VVC and JPEG-AI VM}
\label{BDM_sec}

JPEG-AI VM~3.4 BOP can outperform the classical VVC intra codec by more than 10\% gain in Bjøntegaard delta rate (BD-Rate)~\cite{bjontegaard2001calculation} gain. Researchers attribute this success to the flexible bit distribution in the spatial domain of the NN-based image codec. In contrast, VVC intra's block-wise structure limits its performance. However, based on our observations, we arrived at a different conclusion. This section will introduce the analysis of spatial bit distribution between VVC and JPEG-AI VM~3.4 BOP. To clarify the bit distribution, we utilise the bit distribution map (BDM) which displays the spatial bit distribution in proportion to the original image.   

\subsection{Basic Setting for Bit Distribution Map}

In order to achieve a better comparison of the bit distribution difference between JPEG-AI VM and VVC, we opted for the metric that has the greatest variance between the two codecs. Following our observations, VM and VVC has the largest difference in PSNR of luminance; therefore, we selected images that could produce the greatest disparity between VM and VVC. To ensure consistency, the chosen images were compared at the same target bits per pixel (bpp). However, the algorithm for matching bit rates cannot precisely provide the same bpp. The difference between the target bpp and the actual bpp can only be tolerated to less than 10\%. 

\begin{table*}[t]
	\centering
	\footnotesize
	\begin{tabular}{c|ccccccccccccccccc}
		\toprule
		\textbf{Q index}&\textbf{-8}&\textbf{-7}&\textbf{-6}&\textbf{-5}&\textbf{-4}&\textbf{-3}&\textbf{-2}&\textbf{-1}&\textbf{0}&\textbf{1}&\textbf{2}&\textbf{3}&\textbf{4}&\textbf{5}&\textbf{6}&\textbf{7}&\textbf{8}\\
		\midrule
		\makecell[t]{\textbf{Q step}}
		
		&\makecell[t]{0.25}
		
		&\makecell[t]{0.297}
		
		&\makecell[t]{0.354}
		
		&\makecell[t]{0.420}
		
		&\makecell[t]{0.5}
		
		&\makecell[t]{0.595}
		
		&\makecell[t]{0.707}
		
		&\makecell[t]{0.841}
		
		&\makecell[t]{1}
		
		&\makecell[t]{1.189}
		&\makecell[t]{1.414}
		&\makecell[t]{1.682}
		&\makecell[t]{2}
		&\makecell[t]{2.378}
		&\makecell[t]{2.828}
		&\makecell[t]{3.364}
		&\makecell[t]{4}
		\\
		\bottomrule
	\end{tabular}
	\caption{Quantization index and its corresponding quantization step }
	\label{tab2}
\end{table*}

\begin{figure*}[t]
	\centerline{\includegraphics[width=0.8\linewidth]{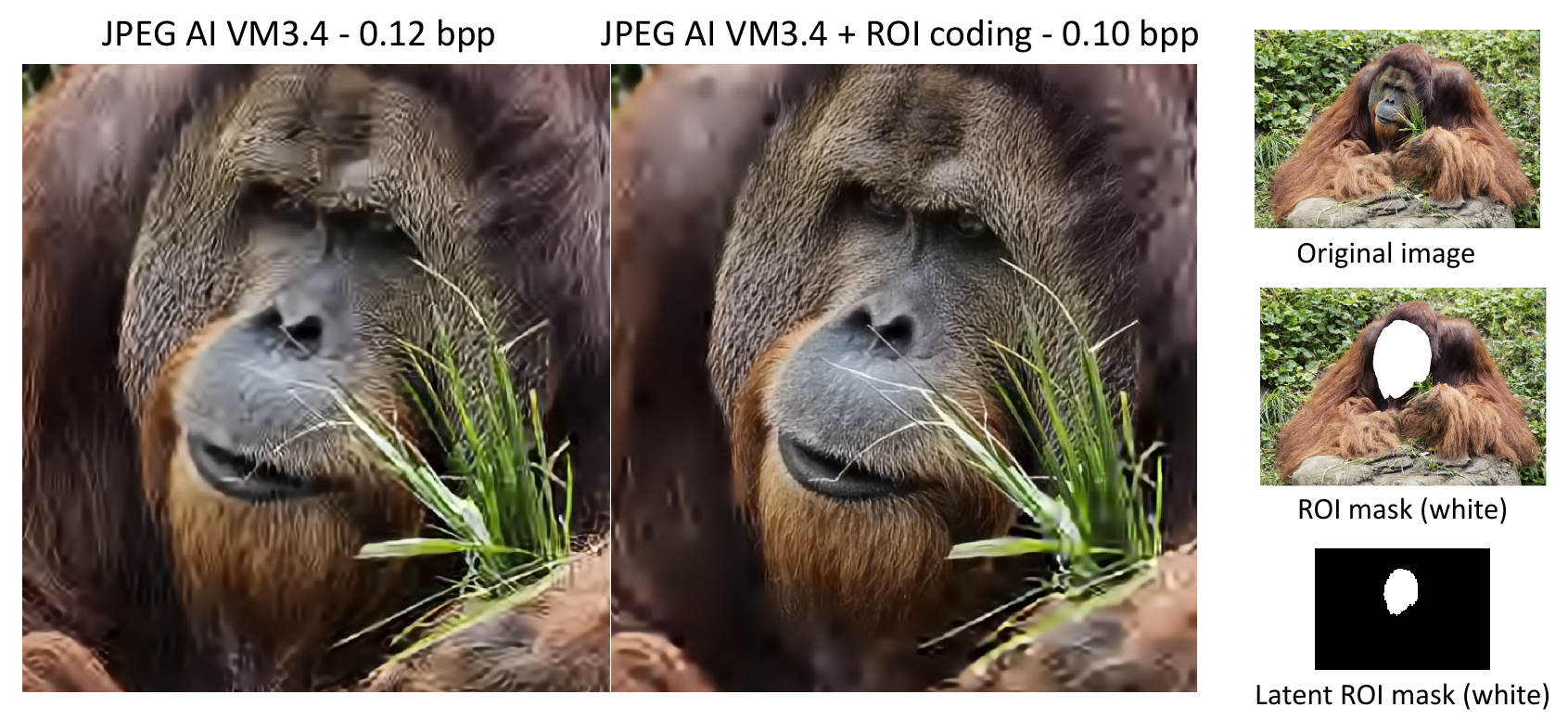}}
	\caption{Example of ROI use case for spatial quality map}
	\label{ROIxing}
\end{figure*}

In JPEG-AI VM, the cost of bits is determined by utilizing the latent and hyper latent tensors, by taking into account only primary component (luminance). In the spatial domain of JPEG-AI VM, each element in the latent tensor represents one $16 \times 16$ block within the initial image, and each element in the hyper latent space represents one $32 \times 32$ block. This is due to the four down-sampled layers in the analysis transform net and one down-sampled layer in the hyper encoder net. To display the bit distribution in the original image scale, the maximum resolution of BDM we can attain is per $16 \times 16$ block.

In contrast, VVC encodes images on a block-by-block basis, dividing them into various shapes and sizes prior to compression. To illustrate how bits are distributed, blocks with different shapes and sizes are utilized in BDM at the original image scale.  

To visualize the BDM of VVC and JPEG-AI VM over a common range of values, we normalized the average bits of each block, denoted by $B$, to the interval $ [0, {\rm max}({\rm max}(B_{VVC}),{\rm max}(B_{VM}))]$. Here, ${\rm max}(B_{VVC})$ denotes the maximum bits per block of VVC, while ${\rm max}(B_{VM})$ refers to the maximum bits per block of JPEG-AI VM. Regarding the lower limit of the interval, we elected to use 0 rather than the minimum bits per block of the two codecs. This decision was made because 0 represents both the theoretical minimum and optimal bit cost. In the resulting BDM, bright pixels indicate that the codec expends more bits on a given area, whereas dark pixels indicate that it consumes fewer bits.

Fig. \ref{fig:BDM_05} shows an example of BDM generation for two codecs.  Fig. \ref{fig:BDM_05_i} displays the original image, while Figures \ref{fig:BDM_05_a} and \ref{fig:BDM_05_b} show the original BDM of JPEG-AI VM and VVC, respectively. It is evident that the BDM of VVC is more complex owing to the variational block size. In contrast, the BDM of JPEG-AI VM has a fixed block size. To compare the bpp of VVC and JPEG-AI VM at the same resolution, as illustrated in Fig. \ref{fig:BDM_05_c}, we also present the VVC's BDM regrouped into $16 \times 16$ blocks. When obtaining the BDM of two codecs at the same resolution, we assess the flexibility of bit distribution by computing the variance of the BDM.

\subsection{Bit Distribution Map Analysis}

Table \ref{tab1} displays the variance of BDM between JPEG-AI VM and VVC for six specific examples, each at a particular bpp point.   These examples were chosen because they present significant PSNR-Y gap between the two codecs. The left side of the table shows three examples where VVC achieved a superior PSNR-Y, whereas the right side shows three examples where VVC resulted in a lower PSNR-Y.

From these examples, it is apparent that the BDM of JPEG-AI VM consistently displays a more uniform spatial bit distribution, with its variance always lower than that of VVC. Therefore, the BD-rate gain of JPEG-AI VM over VVC cannot be attributed to a more flexible bit distribution. Nevertheless, it is generally understood that a codec with a more flexible bit distribution can allocate more bits to vital sections such as the region of interest (ROI). In the upcoming section, the proposed spatial bit distribution tool will be presented.

%% file: sections/QPM.tex
\section{Spatial Quality Map}
To facilitate the spatial distribution of bits in the JPEG-AI VM, a spatial quality map was devised for the codec. This attribute enables different quantization steps to be allocated to diverse parts of the image. In the subsequent section, we present the technical details and experimental results of the spatial quality map.

\begin{table*}[t]
	\centering
	\footnotesize
	\begin{tabular}{c|ccc|ccc}
		\toprule
		\textbf{PSNR-Y of}&\textbf{IMG05 0.75bpp}&\textbf{IMG34 0.75bpp }&\textbf{IMG43 0.75bpp}&\textbf{IMG12 0.25bpp}&\textbf{IMG12 0.5bpp}&\textbf{IMG32 0.75bpp}\\
		\midrule
		\makecell[l]{ VVC\\ JPEG-AI VM\\ JPEG-AI VM + Q map}
		
		&\makecell[l]{46.73\\39.80\\39.86$_{+ 0.06 {\rm dB}}$}
		
		&\makecell[l]{38.33\\33.49\\33.59$_{+ 0.1 {\rm dB}}$}
		
		&\makecell[l]{46.61\\41.61\\41.63$_{+ 0.02 {\rm dB}}$}
		
		&\makecell[l]{32.67\\33.24\\33.69$_{+ 0.45 {\rm dB}}$}
		
		&\makecell[l]{36.79\\37.12\\37.32$_{+ 0.2 {\rm dB}}$}
		
		&\makecell[l]{34.93\\35.99\\36.32$_{+ 0.33 {\rm dB}}$}
		
		\\
		\bottomrule
	\end{tabular}
	\caption{PSNR Y (in dB) of VVC, JPEG-AI VM, and JPEG-AI VM with using quality index map }
	\label{tabpsnr}
\end{table*}

\begin{table*}[t]
	\centering
	\footnotesize
	\begin{tabular}{c|ccc|ccc}
		\toprule
		\textbf{bpp of}&\textbf{IMG05 0.75bpp}&\textbf{IMG34 0.75bpp }&\textbf{IMG43 0.75bpp}&\textbf{IMG12 0.25bpp}&\textbf{IMG12 0.5bpp}&\textbf{IMG32 0.75bpp}\\
		\midrule
		\makecell[l]{ VVC\\ JPEG-AI VM\\ JPEG-AI VM + Q map}
		
		&\makecell[l]{0.760\\0.754\\0.741$_{- 1.7\%}$}
		
		&\makecell[l]{0.778\\0.759\\0.785$_{+ 3.4\%}$}
		
		&\makecell[l]{0.809\\0.743\\0.725$_{- 2.4\%}$}
		
		&\makecell[l]{0.255\\0.254\\0.272$_{+ 7.1\%}$}
		
		&\makecell[l]{0.533\\0.508\\0.530$_{+ 4.3\%}$}
		
		&\makecell[l]{0.793\\0.751\\0.774$_{+ 3.1\%}$}
		
		\\
		\bottomrule
	\end{tabular}
	\caption{Bpp of VVC, JPEG-AI VM, and JPEG-AI VM with using quality index map. (The actual bpp is within 10\% difference from the target bpp.) }
	\label{tabbpp}
\end{table*}

\subsection{Spatial Quality Index Map}
\label{roi_sec}

The input image $x$ in the JPEG-AI VM, with dimensions $(3, H, W)$, is compressed into a latent tensor $y$ with dimensions $(C, \frac{H}{16}, \frac{W}{16})$. $H$ and $W$ are the height and width of the input image, respectively, while $C$ denotes the channel number in the latent domain. One element in the latent tensor represents each $16\times16$ block in the original image spatial scale. For supporting the spatial bit distribution, we propose a spatial quality index map $Q$ with dimensions $(1,\frac{H}{16}, \frac{W}{16})$. Multiplying the latent tensor with $Q$ yields a block-wise distribution of the original image.

When decoding the latent tensor to reconstruct the image, the decoder requires knowledge of $Q$ for inverse quantization. Therefore, $Q$ needs to be signaled to the decoder. However, signaling $Q$ to the decoder results in additional bits in the bitstream. To conserve the signaled bits, we propose signaling the integer quantization index in $Q$, where each quantization index corresponds to a floating quantization step $Q_s$, determined by $Q_s = 2^{\frac{Q}{4}}$. A table is included for reference. Table. \ref{tab2} displays the 17 quantization index values alongside their corresponding quantization steps. Increasing the quantization index results in the codec spending more bits on the relevant area. Positive indices increase the number of bits allocated, while negative indices decrease it. An index of 0 does not affect the latent element. When signaling $Q$, the first step is to obtain its prediction $Q_{pred}$. Following this, we encode the difference $\Delta_{Q} = Q - Q_{pred}$ using the entropy model. The prediction $Q_{pred}$, is calculated based on two adjacent elements of $Q$, as demonstrated in Eq. \ref{eq1}, where $i\in[0,\frac{H}{16}]$ and $j\in[0,\frac{W}{16}]$.
\begin{equation}
    Q_{pred}[i,j] = (Q[i,j-1] - Q[i-1,j])/2   \label{eq1}
\end{equation}

\subsection{ROI Experiment Result}
One application of the spatial quality map is ROI-based coding, where better visual quality can be achieved by assigning more bits to the ROI. This can be followed by reducing bits distributed to the background area for further optimization.    

Fig. \ref{ROIxing} shows an example of ROI coding outcomes for image 21 in the JPEG-AI CTTC test dataset, conducted by the same model. Initially, we selected the ROI mask that targets the ape's face in the original image. We then generated a latent ROI mask, which is a binary quality index map that assigns a quality index of 6 to white areas and a quality index of -6 to black areas. Applying the latent ROI mask facilitates the codec's preservation of the details in the ape's face region. As demonstrated in Fig. \ref{ROIxing}, the utilization of ROI coding in JPEG-AI VM3.4 results in a clearer texture on the ape's face compared to without the ROI coding. Furthermore, to reduce the number of bits used, the background of the image becomes blurred, leading to a 17\% bit savings from 0.12 bpp to 0.10 bpp.

\subsection{Objective Experiment Result}
Section \ref{BDM_sec} analyzed the spatial bit distribution of VVC and JPEG-AI VM by creating the corresponding BDM. The study concluded that VVC possesses a more flexible spatial bit distribution due to its higher variance of the BDM. To further demonstrate that the objective quality can be enhanced by utilizing a more flexible bit distribution, we opted to utilize the BDM of VVC as the quality index map of JPEG-AI VM.

\begin{figure}[t]
	\centerline{\includegraphics[width=0.95\linewidth]{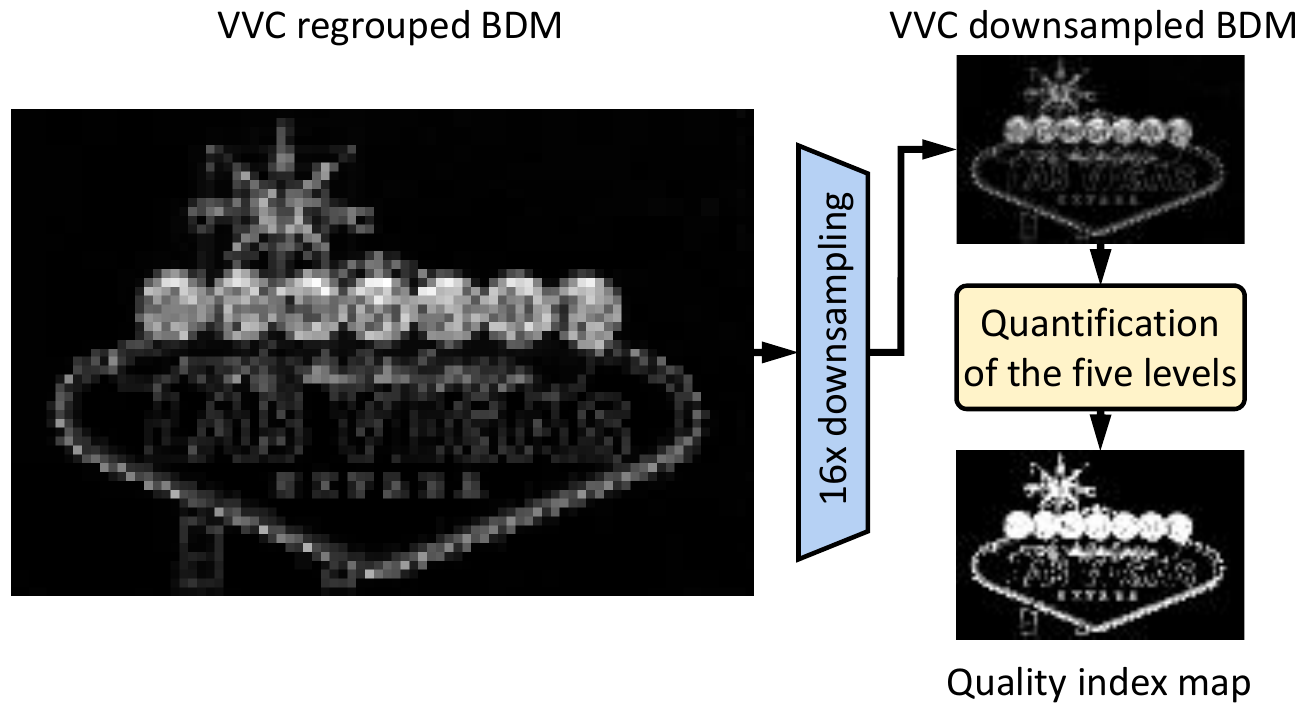}}
	\caption{The process of creating a quality index map from the BDM of VVC}
	\label{indexmap}
\end{figure}

Fig. \ref{indexmap} depicts the process of creating a quality index map from the BDM of VVC.  Initially, we utilized the regrouped $16 \times 16$ block BDM of VVC, which has the same size as the original image $x$. Then, we down-sampled the BDM of VVC 16 times to obtain a down-sampled BDM of VVC with the same size as the latent tenor $y$. We then calculated the mean value $m_{VVC}$ of the down-sampled VVC BDM by $ m_{VVC} = mean(B’_{VVC})$, where $B’_{VVC}$ denotes the average bits of each block in the down-sampled BDM of VVC. Then we quantified the BDM into five levels by comparing it with the $m_{VVC}$, each level indicates a different quality index in Table \ref{tab2}. For each element $B’_{VVC}$ in the down-sampled BDM, we assign a quality index based on its value: if $B’_{VVC}$ is within the range $[0, m_{VVC})$, the corresponding quality index is -1; if $B’_{VVC}$ is greater than or equal to $m_{VVC}$ and less than $1.5m_{VVC}$, the quality index is 0; if $B’_{VVC}$ is within the range $[1.5m_{VVC}, 2.5m_{VVC})$, the quality index is 1; if $B’_{VVC}$ is within the range $[2.5m_{VVC}, 4m_{VVC})$, the quality index is 2; and finally, if $B’_{VVC}$ is greater than or equal to $4m_{VVC}$, the quality index is 3. By utilizing a five-level quality index, we allocate more bits to the areas where VVC also spends more bits. Additionally, we employ index -1 to decrease the number of bits utilized for less significant areas in order to limit the total number of bits used.

The PSNR-Y results obtained from implementing the VVC BDM-based quality index map on the JPEG-AI VM are presented in Table \ref{tabpsnr}. The use of the quality index map substantially improves all cases for the selected images in Section \ref{BDM_sec}. Notably, the PSNR-Y for the image 12 at 0.25 bpp increased from 33.24 dB to 33.69 dB, representing an increase of nearly 1.4\%. Furthermore, the use of the quality map index -1 facilitated achieving the target bpp with minimal deviation. Table \ref{tabbpp} illustrates that the highest increase in bpp is only 7.1\% (from 0.255 bpp to 0.272 bpp), for image 12 at 0.25 bpp. Interestingly, in certain instances, like image 43 at 0.75bpp, the observed bpp value decreased from 0.743 bpp to 0.725 bpp, resulting in a 2.4\% reduction in bits.

In summary, utilizing the VVC BDM-based quality index map on the JPEG-AI VM can result in a greater PSNR-Y gain, suggesting that a more adaptable spatial bit allocation can enhance NN-based image compression codecs. However, directly utilizing VVC BDM in JPEG-AI VM demands excessive complexity from the VVC encoder. Therefore, quality enhancement of JPEG-AI VM by this approach is impractical. The experiment performed in this subsection only shows the potential of quality enhancement of JPEG-AI VM by using a more adaptive spatial bit distribution.

%% file: sections/Conlusion.tex
\section{Conclusion}
The purpose of this study was to compare the spatial bits distribution of JPEG-AI VM and VVC by generating and comparing their bit distribution maps. Surprisingly, our findings show that VVC has a more flexible bit distribution due to the variation in block shape and size. In contrast, the JPEG-AI VM is a NN-based image codec, and it operates at picture level and its design is comparable with classical codec which uses fixed block size.

In order to release this limitation and enable better spatial bits allocation for learnable codec, we propose a spatial quality index map to increase the flexibility of the bit distribution in JPEG-AI VM. Our experiments show that the quality index map not only saves bits, but also improves visual quality. It is worth noting that the VVC encoder utilizes a complex strategy involving multiple cost calculations and comparisons to determine optimal block splits and bit allocations. By utilizing the VVC bit distribution strategy, JPEG-AI VM's objective performance can be further improved.